\documentclass[conference,twocolumn,transmag]{IEEEtran}
\usepackage{graphicx}
\usepackage{subcaption}
\usepackage{hyperref}
\usepackage{cite}
\usepackage{amsmath, amsfonts, amssymb, textcomp}
\usepackage{amssymb}
\usepackage{algorithm}
\usepackage{flushend}
\usepackage[noend]{algpseudocode}

\makeatletter
\def\algbackskip{\hskip-\ALG@thistlm}
\makeatother

\begin{document}
\title{A Proposed Artificial intelligence Model for Real-Time Human
Action Localization and Tracking}

%\author{\IEEEauthorblockN{Ahmed Ali Hammam}
%\IEEEauthorblockA{Faculty of Computers and Information\\
%Cairo University\\
%Scientific Research Group in Egypt\\ (SRGE)\\
%ahmed.a.hammam@grad.fci-cu.edu.eg\\
%www.egyptscience.net}
%\and
%\IEEEauthorblockN{Mona M.Soliman}
%\IEEEauthorblockA{Faculty of Computers and Information\\
%Cairo University\\
%Scientific Research Group in Egypt\\ (SRGE)\\
%Mona.solyman@fci-cu.edu.eg\\
%www.egyptscience.net}
%\and
%\IEEEauthorblockN{Aboul Ella Hassanein}
%\IEEEauthorblockA{Faculty of Computers and Information\\
%Cairo University\\
%Scientific Research Group in Egypt\\ (SRGE)\\
%aboitcairo@gmail.com \\
%www.egyptscience.net}}

\author{\IEEEauthorblockN{Ahmed Ali Hammam\IEEEauthorrefmark{1}\IEEEauthorrefmark{,2},
Mona Soliman\IEEEauthorrefmark{1}\IEEEauthorrefmark{,2},and
Aboul Ella Hassanien \IEEEauthorrefmark{1}\IEEEauthorrefmark{,2}}
\IEEEauthorblockA{\IEEEauthorrefmark{1}Faculty Of Computer And Information,
Cairo University, Egypt}
\IEEEauthorblockA{\IEEEauthorrefmark{2}Scientific Research Group in Egypt,SRGE, Egypt}}

\IEEEtitleabstractindextext{%
\begin{abstract}
\textit{Abstract-} In recent years, artificial intelligence (AI) based on deep learning (DL) has sparked tremendous global interest. DL is widely used today and has expanded into various interesting areas. It is becoming more popular in cross-subject research, such as studies of smart city systems, which combine computer science with engineering applications. Human action detection is one of these areas. Human action detection is an interesting challenge due to its stringent requirements in terms of computing speed  and accuracy. High-accuracy real-time object tracking is also considered a significant challenge. This paper integrates the YOLO detection network, which is considered a state-of-the-art tool for real-time object detection, with motion vectors and the Coyote Optimization Algorithm (COA) to construct a real-time human action localization and tracking system. The proposed system starts with the extraction of motion information from a compressed video stream and the extraction of appearance information from RGB frames using an object detector. Then, a fusion step between the two streams is performed, and the results are fed into the proposed action tracking model. The COA is used in object tracking due to its accuracy and fast convergence. The basic foundation of the proposed model is the utilization of motion vectors, which already exist in a compressed video bit stream and provide sufficient information to improve the localization of the target action without requiring high consumption of computational resources compared with other popular methods of extracting motion information, such as optical flows. This advantage allows the proposed approach to be implemented in challenging environments where the computational resources are limited, such as Internet of Things (IoT) systems. The experimental results obtained using the proposed model show its superiority with respect to various other online and offline systems in terms of accuracy and calculation time for the detection and tracking of human actions in various video sequences.

\end{abstract}

}
\maketitle

\IEEEdisplaynontitleabstractindextext

\section{Introduction}
\IEEEPARstart{A}{rtificial} Intelligence (AI) is intelligence exhibited by the machine. In computer science, the field of AI research is represented as the study of "intelligent agents": machines that can perceive their environment and act accordingly to maximize their chances of achieving a certain goal \cite{Ongsulee} . Colloquially, the term "artificial intelligence" is used to refer to cases in which a machine imitates "cognitive" functions similar to those of humans. However, until circa 2012, AI research was restricted to sophisticated technology businesses, governments, and research organizations, fueling both perceptions. Since then, AI has broken away from the hypothetical and into real-world company alternatives. Many of these alternatives are motivated by the broad accessibility of graphics processing units (GPUs), which makes parallel processing quicker and cheaper.

AI now includes many sub-fields and relies on a variety of techniques such as neural networks (e.g. brain modeling, time series prediction, classification), evolutionary computation (e.g. genetic algorithms, genetic programming), swarm intelligence (e.g. particle swarm optimization), machine vision (e.g. object recognition, image comprehension), Robotics (e.g., smart control, independent exploration), specialist systems (e.g. decision support systems, teaching systems), voice processing (e.g. voice recognition and manufacturing), natural language processing (e.g. machine translation), planning (e.g. scheduling, game play) and machine learning (e.g. decision tree learning, space learning version) \cite{Russell}. Most of these fields and techniques have aspects of both science and engineering.

Historically, researchers have turned to nature for guidance when designing AI systems. Not surprisingly, the first model to be explored was the most familiar to our own brains. Starting with the perceptron’s of the 1950s and continuing to this day, neural networks and other neurologically inspired architectures are the dominant models for AI research. Recently, neural-based deep learning (DL) has become one of the preeminent machine learning techniques based on information learning depictions. DL (also known as deep structured learning, hierarchical learning or profound machine learning) is the study of artificial neural networks and associated machine learning algorithms containing more than one hidden layer \cite{Michelucci}. There are many layers between input and output in a profound network (and these layers are not made up of neurons, although it may assist to think about them that way). In DL \cite{LeCun}, a computer model learns straight from pictures, text, or sounds to execute classification functions. DL models can attain state-of-the-art precision, sometimes surpassing human output. Models are taught using big sets of labeled information and architectures of neural networks that contain many layers. Various architectures of deep learning, such as profound neural networks, convolutionary profound neural networks, profound networks of beliefs and recurrent neural networks.

Beside Neural network, There is another sub-field of nature-inspired AI, namely swarm intelligence. Billions of years of evolution have generated at least one alternative technique of constructing high-level intelligence that is not neural-rather collective \cite{Parpinelli}.

Swarm intelligence methods are based on the research of collective behavior in distributed systems. Such a system is made up of a population of simple agents communicating locally to each other and to their environment\cite{Panigrahi}. The scheme is initialized with an individual population (i.e. prospective solutions). These individuals are then modified through several phases of repetition by imitating the social behavior of insects or animals in an attempt to find the optimum in problem space. Exploring the search space is enhanced by changing each prospective alternative according to past experiences of the individual and their interactions with other members of the population and with the surroundings.

This work aims to utilize  recent research in various AI sub-fields to provide an automated solution for human action localization and tracking in real-time videos. The problem of automated action understanding is becoming increasingly relevant as the enormous technological developments occurring in daily life are giving rise to a need for end-to-end security and monitoring solutions that can function in challenging environments. Localization of real-time action corresponds to the task of simultaneously locating actions and detecting their classes in real-time from input video streams. Real-time action localization is a challenging problem that requires expensive features that are difficult or impossible to extract due to the real-time processing requirements and device limitations. \par

Advances in localization and tracking of human behavior are linked to advances in study fields such as object identification, human dynamics, domain adaptation and semant segmentation. Over the past century, the associated techniques have developed from early systems, whose applications are often restricted to certain controlled settings and sophisticated solutions that can learn from millions of videos and be implemented to almost all daily operations. Given the wide spectrum of associated apps, from video surveillance to human-computer interaction, science milestones in action localization are being accomplished increasingly quickly, ultimately leading to the disappearance of techniques that used to be efficient within an ever-short timeframe.\par

Enabled by rapid developments in AI and machine learning and by the success that has been achieved using DL approaches in the processing of still images for tasks such as human action recognition, video analysis tasks such as recognition and detection have been evolving from the relatively simple classification of present states to the prediction of future states. The localization and classification of actions must be performed even before the actions are fully observed\cite{Baek2017}. Many successful solutions have been introduced \cite{Singh2017}, \cite{Peng2016}, \cite{Saha2016}, \cite{Shou2016}, \cite{Peng2016} and \cite{Weinzaepfel2015} for both real-time action localization \cite{Singh2017} and offline action localization \cite{Peng2016}, \cite{Saha2016}, \cite{Shou2016}, \cite{Peng2016} and \cite{Weinzaepfel2015}. Most of these solutions follow the two-stream approach, in which there are two input streams, i.e., appearance and motion streams, with appearance information being extracted from RGB frames and motion information being extracted from the optical flows of the input. Such a two-stream architecture can achieve excellent performance for action localization, with high accuracy and speed.\par

This paper presents a new model that enables the detection and tracking of human actions and actors in real time videos. The resource requirements of the proposed approach are relatively inexpensive compared with previous approaches, allowing the proposed model to be implemented in challenging environments where computing resources are limited, such as Internet of Things (IoT) systems. This model can provide a complete solution for surveillance scenarios in which there is a need to detect actions and also track the corresponding actors in an environment with occlusion. This is considered an important issue in security systems.

\textbf{Contributions}. The main contributions of this work can be summarized as follow:
\begin{itemize}
  \item \textbf{First :}Propose a novel real-time action localization model that can be implemented in challenging environments (e.g., IoT systems).
  \item \textbf{Second :}Provide a solution to the challenges of optical flow methods by using motion vectors instead of optical flows to extract motion information while still achieving good accuracy compared with state-of-the-art approaches \cite{Singh2017}\cite{Peng2016}], in which the most expensive step preventing these previous approaches from achieving real-time performance is the calculation of the optical flows.
  \item \textbf{Third :}Ensure that the proposed model can be implemented in challenging environments, such as an IoT environment, in which the available resources and speed are limited compared with DL using powerful CPU and GPU devices.
  \item \textbf{Fourth :}Improve the accuracy of the motion detection network by training an optical flow detection network and then using the weights of this network for transfer learning to a motion-vector-based detection network.
  \item \textbf{Fifth :}Incorporate the Coyote Optimization Algorithm (COA) to track the actor trying to perform every detected action in real-time and introduce a new COA-based tracking model.
\end{itemize}
The rest of the paper is structured as follows. In Section 2, the basics and background related to the proposed approach are first discussed, followed by a discussion of recent state-of-the-art approaches for action detection and localization in Section 3. The proposed system is presented in Section 4, and the results are discussed in Section 5. Section 6 summarizes the main findings of this paper.

\section{Basic and Background}

%                   \subsection{Deep learning in computer vision}

\subsection{Coyote Optimization Algorithm (COA)}
Coyote Optimization Algorithm (COA)\cite{Pierezan2018 } is considered to be a population-based algorithm inspired by the species of Canis latrans classified as swarm intelligence and developmental heuristic and affected by coyote behavior \cite{Pierezan2018}.
In the COA, The coyote population is split into Np $\in$ (N * packs) with Nc $\in$ (N * coyotes) each. The number of coyotes per package in this first proposal is static and similar for all packages. Consequently, Np and Nc multiply the complete population in the algorithm. For simplification purposes, In this first version of the algorithm, solitary (or transient) coyotes are not considered. Each coyote is a feasible answer to the optimization problem and the cost of the objective function is its social situation.
The COA mechanism was intended on the basis of coyote social conditions, which means the decision variables $\vec{X}$ of an global optimization problem. Thus, the social condition soc (set of decision variables) of the cth coyote of the pth pack in the tth instant of time:
\begin{equation}
	\label{EQ:socialCondition}
	soc^{p,t}_{c}=\vec{X}=(X_{1},X_{2},...,X_{D})
\end{equation}
And This means in the coyote’s adaptation to the environment (the objective function cost) \emph{fit cp} \emph{,t $\in$ R}.
initialize the global population of coyotes is the first steps in COA. COA is consider a stochastic algorithm, the initial social conditions for each coyote are determined randomly. This is done by assigning random values within the search space for the cth coyote of the pth pack of the jth dimension :

\begin{equation}
	\label{EQ:coyoteSpace}
	soc^{p,t}_{c,j}=lb_{j}+r_{j}.(ub_{j}-lb_{j})
\end{equation}

where in \emph{lbj and ubj} its representatives, the lower and upper bounds of the $j_{th}$ decision variable respectively, D is the search space dimension and $r_{j}$ is a real random number with range [0,1].the coyotes’ adaptation in the respective current social conditions are evaluated:

\begin{equation}
	\label{EQ:social}
	fit^{p,t}_{c}=f(soc^{p,t}_{c})
\end{equation}
COA only considers one alpha that is best suited for the environment. Considering the problem of minimization, the alpha of the $p_{th}$ pack in the $t_{th}$ instant of time is defined as:
\begin{equation}
	\label{EQ:alphaPack}
	alpha^{p,t}={soc^{p,t}_{c}|arg_c{1,2,...,N_{c} minf(soc^{p,t}_{c})}}
\end{equation}
Because of the obvious indications in this species of swarm intelligence, the COA assumes that the coyotes are organized enough to share and to contribute  the social conditions to the pack’s maintenance. Thus, the COA connects all data from the coyotes and calculates it as a cultural tendency of the pack as the median social conditions of all coyotes from that specific pack:

\begin{equation}
	\label{EQ:median}
	clut^{p,t}_{j} = \begin{cases} O^{p,t}_{\frac{(N_c+1)}{2},j} ,& N_{c}\space is \space odd \\
\frac{O^{p,t}_{\frac{N_c}{2},j}+O^{p,t}_{(\frac{N_c}{2}+1),j}}{2} ,& otherwise \end{cases}
\end{equation}
Where \emph{Op,t} represents the ranked social conditions of all coyotes of the $p_{th}$ pack in the $t_{th}$ instant time for each j in the range [1, D].
To compute the culture influence COA assumes that coyotes are under the alpha influence ($\delta_{1}$)and the pack influence ($\delta_{2}$).
\begin{equation}
	\label{EQ:alpha}
	\delta_{1}=alpha^{p,t}-soc^{p,t}_{cr1}
\end{equation}
\begin{equation}
	\label{EQ:pack}
	\delta_{2} =clut^{p,t}-soc^{p,t}_{cr1}
\end{equation}
Where $soc^{p,t}_{cr1}$ and $soc^{p,t}_{cr2}$ are a random coyote.
the coyote’s new social condition is update by using the alpha and the pack influence using these equation:
\begin{equation}
	\label{EQ:newsocial}
	new\_soc^{p,t}_{c} =soc^{p,t}_{c}+r_{1}.\delta+r_{2}.\delta
\end{equation}
Where $\delta_{1}$ and $\delta_{2}$ are the weights of the alpha and the pack influence , respectively.  r1 and r2 Initially defined as random numbers inside the range [0, 1]. The new social condition is computed with the following :
\begin{equation}
	\label{EQ:newfit}
	new\_fit^{p,t}_{c} =f(new\_soc^{p,t}_{c})
\end{equation}
Coyote’s cognitive capacity decide if the new social condition is better than the older one to keep it, it means:
\begin{equation}
	\label{EQ:cognitive}
	soc^{p,t+1}_{c} = \begin{cases} new\_soc^{p,t}_{c} ,& new\_fit^{p,t}_{c} < fit^{p,t}_{c} \\
soc^{p,t}_{c} ,& otherwise \end{cases}
\end{equation}

\subsection{Motion Vector}

Motion vectors \cite{Wiegand2003 } Originally suggested for video encoding by saving image modifications from one frame to the next. It is intended to use the motion data of the respective image blocks to decrease the video bit rate. We can use these vectors to detect and track motion and replace traditional motion information extractor such as optical flow.\par

The vector of motion is similar to the optical flow. Both are two dimensional vectors to describe the respective data on the motion of pixels in two ongoing frames. Unlike optical flow, vector motion is widely used in various HEVC \cite{Bross2013} and H.264 \cite{Wiegand2003} video coding norms. Motion vector is accessible in a compressed video stream and can be achieved directly with almost no computing expenses. This property makes motion vector an attractive substitute for optical flow to achieve an effective action analysis.\par

There are many video compression standards like H.264 \cite{Wiegand2003} and  HEVC \cite{Bross2013} , the input video frames are coarsely divided into macro-blocks (MBs), which form the basis for inter (and intra) prediction.
Inter-predicted macro-blocks (MBs) are (optionally) partitioned into blocks that are predicted via motion vectors representing the displacement from matching blocks in previous or subsequent frames. MB motion information can be extracted from a compressed video bitstream using different available library.\par

\section{LITERATURE REVIEW}
This section reviews related works, about the problem of action detection and tracking in a video sequence. We explore the main related work into three categories: 1) Action Recognition, 2) Action Localization, and 3)Object tracking
%----------------------------------------------------------------------------------------
%	Action Recognition
%----------------------------------------------------------------------------------------
\subsection{Action Recognition}
In this section, we briefly present the main families of methods for action recognition. Action recognition (classification) can be considered a video classification problem it can be defined as  assign a set of predefined action classes to a video. It is . Action recognition comprises of three major steps consists of three main steps as follow : first step feature extraction, second step a representation for a video based on the extracted features, and finally classification of the video using the representation. There is an amount of work in action recognition with several recent surveys [\cite{Weinland2011}, \cite{Aggarwal2011}].\par
In \cite{Zhang2016} They Provide a realtime action recognition method with high-performance and accuracy. Their approaches accelerate the two stream CNN architecture to speed reach to 390.7 FPS, but their approach deals with action recognition problem and not deal with action localization problem.\par
In \cite{Gammulle2017} They deal with the issue of human action recognition from sequences of videos. Motivated by the exemplary results obtained via deep learning and automatic feature learning approaches in computer vision, They focus their work towards learning salient-spatial features via a convolutional-neural-network (CNN) and then map their temporal relationship by use Long Short Term Memory ( LSTM ) networks.\par

%----------------------------------------------------------------------------------------
%	Action Localization
%----------------------------------------------------------------------------------------
\subsection{Action Localization}
Action localization, called also action detection, refers to the problem
of recognizing the actions Including their extent. In this paper, we concentrate on
human action localization in time and space. Also we give a consider actions performed by animals a perspective view point. Significant attention was given to Action Localization in the last few years [\cite{Jain2015}, \cite{Wang2014}]. Action localization focuses on detecting the actions inside the videos. Action localization was the objective of less studies than Action recognition. Action localization problem is much more advanced than action recognition. Successful action localization requires the action class to be correctly recognized, and also its spatio-temporal location to be identified. In action localization, we can consider action recognition as a sub-problem of it.\par
In this section, we review most techniques for online and offline action localization.
%----------------------------------------------------------------------------------------
%	Offline Action Localization
%----------------------------------------------------------------------------------------
\subsubsection{Offline Action Localization}
In offline mode we have the full video we can know sub-actions, we deal with all frame and processing time, not a challenge. In \cite{Gkioxari2015} introduce models that can be located and classified the actions in the video using convolutional-neural-networks on kinematic and static  cues. Using motion-saliency to eliminate regions that are unlikely to include the action. They extract spatiotemporal feature representations to build strong classifiers using Convolutional-Neural-Networks. They train two Convolutional-Neural-Networks for Action detection task, Motion CNN and Spatial CNN closer to RCNN \cite{Girshick2014}. Spatial-CNN captures the actor's appearance and the cues from the scene. Motion CNN, capture motion Patterns operates on the optical flow and the movement of the actor. And finally, they make the prediction after the use of specially trained SVM classifiers.The SVM classifiers trained on the spatio temporal representations produced by both CNNs. They test their method on UCF Sports \cite{Rodriguez2008} dataset and J-HMDB \cite{Jhuang2013} dataset. on UCF sports, achieving a threshold = 0.6 , with an improvement of 87.3\%, and have a Mean AUC = 41.2\% in comparison with to other approaches .on J-HMDB a larger dataset they can accomplish an accuracy of 62.5\%. In \cite{Jain2014} introduce a strategy to produce 2D+t se-quences of bounding boxes, called tubelets. They have two contributions first, using super-voxels instead Of super-pixels to produce spatio-temporal shapes. Second, to identify the action motion from the background Motion they use independent motion evidence as a feature. They test their approach on two datasets UCF Sports \cite{Rodriguez2008} and MSR-II \cite{Cao2014}. On UCF they obtain 80.24\% of accuracy. On MSR-II tubelets significantly outperform 46.0 \% for Boxing, 31.4 \% for Handclapping and 85.8 \% for Hand-waving. In \cite{Kong2014} this research aims at predicting the action class of a partly observed video before the action is end. Develop a new formulation of learning to capture temporal evolution. For the early recognition of incomplete actions they propose a new multiple-temporal-scale-support-vector-machine (MTSSVM) formulated based on the structured SVM, They test the proposed MTSSVM approach on three datasets: the UTInteraction dataset (UTI) Set 1 (UTI $\sharp$1) and Set 2 (UTI $\sharp$2) \cite{Ryoo2010}, and the BIT Interaction dataset (BIT) \cite{Kong2012}.on the UTI $\sharp$1 dataset their approach obtain a recognition accuracy 78.33\%  only half frames of test videos will be noted. And 95\% recognition Accuracy with full videos. On the UTI $\sharp$2 datasets, their MTSSVM achieves 75\% and 83.33\% prediction results. 75\% accuracy when only half frames of test videos will be noted. And an 83.33\% recognition Accuracy with full videos. On the BIT-Interaction dataset .their method achieves 60.16\% recognition accuracy with only the first 50\% frames of testing videos are observed. In \cite{Heilbron2016} introduce a proposed method which aims to obtain temporal segments actions from untrimmed Videos. Propose a framework for scoring temporal segments according to how likely they are to contain an action. Test their method into a standard activity detection framework. Using two datasets The MSR-II Action dataset \cite{Cao2014} and THUMOS 2014 Detection Challenge dataset \cite{Jiang2014}. On the MSR-II achieve a 60.3, compared to 54.5\% obtained by APT \cite{Gemert2015}.On the Thumos14 detection achieve a 13.5\%, compared to 14.3\% obtained by the top per-former in Thu-mos14.
%----------------------------------------------------------------------------------------
%	Online Action Localization
%----------------------------------------------------------------------------------------
\subsubsection{Online Action Localization}
In online action localization, action localization proplem is becoming an action detection and prediction problem. it deals with the video stream as frame by frame and here processing time is considered a chal-lenge due to real-time constraints. A little work is done here there is more to do here. In \cite{Soomro2016} introduced Online Action Localization in a streaming video to localize and pre-dict actions in an online manner. Using pose-estimation to learn
a mid-level super-pixel based foreground model at every instant. Using dynamic programming on SVM to predict the confidences and label for action segments.
Test their approach on two datasets, JHMDB \cite{Jhuang2013} UCF Sports \cite{Rodriguez2008}.They test the observed video for various overlaps thresholds ( 10\%-60\% ) for JHMDB and UCF Sports. for the JHMDB dataset, Initially, the results enhance but then deteriorate at 60 \% observation percentage. For UCF Sports localization improves and then worsens unexpectedly at 15\% observation percentage. In \cite{Sharaf2015} provide a real-time system for action detection. Build method to detect human-actions in 3D-skeleton-sequences. Their system extracted features from 3D-skeleton-data. They perform action detection using a Support-Vector-Machine (SVM) classifier with a linear kernel. Feature selection is per-formed using the Recursive-Feature-Elimination algorithm (SVM-RFE). they test their result on two datasets: MSRC-12 \cite{Bloom2012}, and G3D \cite{Sharaf2015}. On MSRC-12 improvements reached up to 7.7\%. 16.5\%, for “image and text” and 8.8\% for “text” modality.
On G3D Results for the “Fighting” action is reached to 0.937. In \cite{Baek2017} introduce a random-forest (RF)-based online action detection framework. They using a convolutional-neural-network (CNN)-based features obtained from the RGBD raw images and the relationships between the temporal context present in the past and future frames. Using random forests (RF) algorithm where the refined RF parameters are learned with the aid of contexts. They test their result on using three datasets MSRAction3D \cite{Li2010}, G3D datasets, and OAD \cite{Li2016}. On MSR Action3D Dataset using the unsegment setting obtained maximum 6\% improvement compared to other methods. On G3d their ’RF’ baseline is less competitive than compared approaches. On OAD the performance of their ’RF’ baseline is similar to compared approaches. In \cite{Singh2017} a huge move have been achieved there are successfully integrate SSD \cite{Liu2016} in their approach provide the best speed and accuracy compared with other approaches.

\subsection{Object Tracking}
Object tracking is an optimization method for estimating a target's locations and movement in a video sequence where the first frame is given the target's initialized position.
This method of optimization can be classified as either stochastic or deterministic methods.Deterministic method example, Snakes model \cite{Seo2005} , Mean-shift [\cite{Hu2008},\cite{Li2008}] and Trust region \cite{Liu2004}, are iteratively searching for ideal solution for the objective function. Stochastic method example,  Kalman-Filter (KF) [cite{Reid1979},\cite{Maybeck1990},\cite{Cuevas2005} and \cite{Huang2002}] and Particle-Filter (PF) [\cite{Arulampalam2002},\cite{Yang2005} and \cite{Cho2006}], they are generally fast, but in many real-world object monitoring systems their efficiency is restricted. For example, -Filter (KF) can not deal with a non-Gussian or non-linear problem and Particle-Filter (PF) tends to suffer from heavy computing costs because it requires a higher number of particles to represent the posterior volume of the object state.

Over the last few years, Particle-Swarm-Optimization (PSO) many researchers have used it in object tracking to overcome the limitations of conventional KF and PF in object tracking. Particle-Swarm-Optimization (PSO), first method by \cite{Kennedy2006}, is a population-based stochastic heuristic algorithm designed to tackle complex optimization problems. It is motivated by the conduct of natural swarms, such as bird flocking and fish education.In \cite{Anton2006} Applied PSO in object tracking by pixel-flying particles where tracking of objects arose from the interaction between particles and their environment. In \cite{Zhang2008} implement a sequential particle-swarm-optimization strategy by integrating temporal continuity data into PSO.In \cite{Sha2015}, implement a classified PSO algorithm that applies different search strategies on particles based on their fitness values. Overall, owing to its rapid exploration capacities, PSO is an efficient way to track items. Early convergence, however, is an important issue with PSO. Once particles converge prematurely into any specific region of the search space, the entire swarm stagnates. This will have poor results, especially if there is partial or complete occlusion in video sequences due to absence of efficiency for sub-optimal alternatives.Some variants of PSOs like \cite{Sha2015 } Adding one or more parts to enhance monitoring efficiency for PSO equations. However, this increases the complexity of PSO, so the computation price is generally high.

\section{Proposed Method of Human Action localization and tracking}
The localization of human actions of interest across multiple frames and the tracking of the actors performing these actions are still challenging tasks. To this end, the proposed model attempts to satisfy all requirements of the real-time human action localization and tracking problems and to ensure that any human action captured in a video will be localized and that the corresponding actor will be successfully tracked. Additionally, we attempt to improve the convenience of the model with respect to IoT system requirements. With this model, we introduce a fully real-time approach for human action localization and tracking in videos based on learning to directly detect an action and predict its class. Our input is a video that consists of a sequence of frames captured over time; we will process this video frame by frame. We can think of our video as a volume of data in which each data frame is a function of space and time.

\subsection{Action Localization and Tracking Model}
For the first phase of real-time human action localization and tracking, we adopt a two-stream architecture, as shown in Figure 1. The proposed real-time human action localization model consists of a video decoder and a two-stream  architecture (in which each
stream contains a YOLO \cite{Redmon2017} or SSD \cite{Soomro2012} detection networks). The video decoder takes input from a compressed video stream as it directly collects RGB and motion vector frames during the decoding stage.The RGB and motion vector frames are then fed in two distinct streams to perform real-time action localization and prediction for each frame. The primary distinction between our suggested technique of localization of human action in real time and other methods \cite{Singh2017}, \cite{Peng2016}, \cite{Saha2016}, \cite{Shou2016}, \cite{Peng2016} and \cite{Weinzaepfel2015} is that during these steps, our technique does not involve optical flow computations. This allow our model to be more suitable for usage in an IoT setting.

 Additionally, YOLO \cite{Redmon2017} or SSD \cite{Soomro2012} Are used in our framework to obtain high-level motion and appearance data, respectively. Optical flow computations, which are considered the most time-consuming part of past techniques, are presented in our real-time action localization model, making it more convenient to execute in real-time situations.

\begin{figure*}
  \centering
  \includegraphics[width=\textwidth]{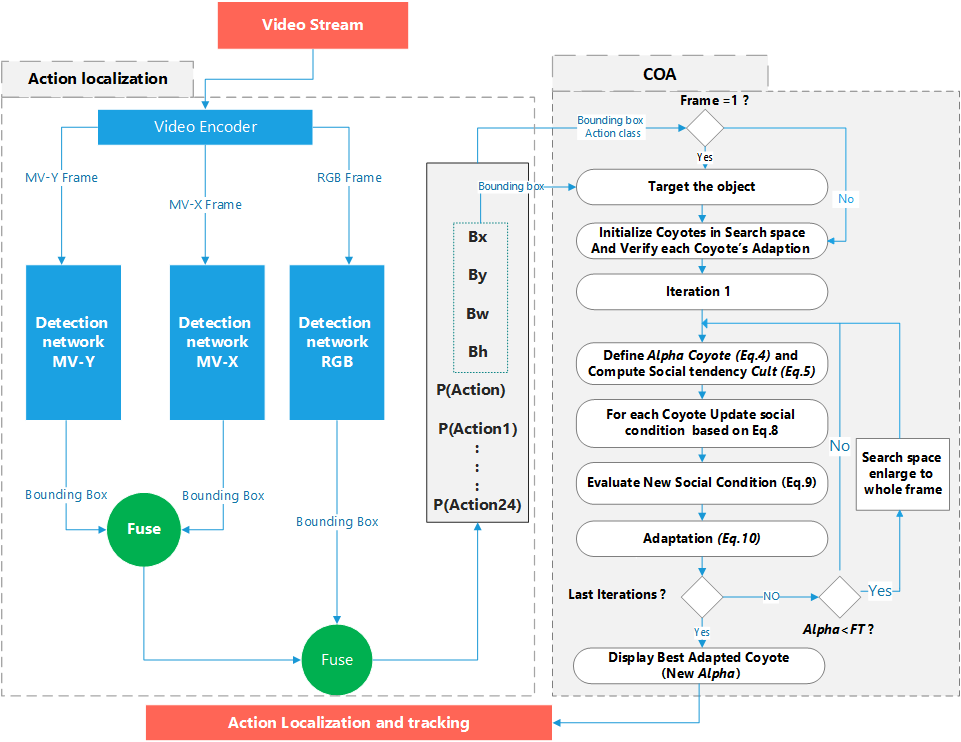}
  \caption{Real time human action localization and tracking model Using Motion Vector.}
  \label{fig:COAModel}
\end{figure*}

First, we extract the current RGB frame and the current motion vector (x,y) frames and input each frame into the corresponding  suitable network that has been trained on test data of the same form.  Each detection network outputs the class of the detected action and its bounding box, if found. We take the information related to the action location, which is the current location of the actor, and input it into the human tracking model (a COA-based tracker), which uses this location Information to extract the target actor's features to be tracked and start the tracking process.

\begin{algorithm}
    \caption{Real Time Action localizations and Tracking algorithm}\label{alg:COAModel}
    \hspace*{\algorithmicindent} \textbf{Input} :  Frame \( \rightarrow\text{   Input Stream of RGB Frame}\)\\
    \hspace*{\algorithmicindent} \textbf{Output} : \(\text{$CA$} \rightarrow \text{Class of Action by Appearance Stream }\)\\
    \hspace*{\algorithmicindent} \textbf{Output} : \(\text{$CM $} \rightarrow \text{Class of Action by  Motion Stream }\)\\
    \hspace*{\algorithmicindent} \textbf{Output} : \(\text{$BA$}\rightarrow \text{Boxes From Appearance Stream}\)\\
    \hspace*{\algorithmicindent} \textbf{Output} :\(\text{$BM$} \rightarrow \text{Boxes From Motion Stream}\)\\
    \hspace*{\algorithmicindent} \textbf{Output} :\(\text{$FB$} \rightarrow \text{Final Boxes From Fused Method}\)\\
    \hspace*{\algorithmicindent} \textbf{Output} :\(\text{$BTrack$} \rightarrow \text{Final Boxes from COA Track Method}\)\\
    \begin{algorithmic}
    \While{hasFrame ()} \par
           \BState Extract Motion vector and RGB Frame By Video Decoder\par
            $Frame_{MVX},Frame_{MVY},Frame_{RGB}, \gets Video_Decoder($Frame$) $ \par
           \BState Input the RGB Frame on Appearance Detection network and extract Bounding Boxes BA and Class of Action CA for appearance stream.\par
            $ $BA$,$CA$ \gets Detection_{Appearance}(Frame_{RGB})$\par
           \BState Input the Motion vector Frame on the other Detection Net-works and extract Bounding Boxes BM and Class of Action CM for motion stream.\par
            $ $BMVX$,$CM$ \gets Detection_{MVX}(Frame_{MVX})$\par
            $ $BMVy$,$CM$ \gets Detection_{MVY}(Frame_{MVY})$\par

           \BState Take the Output of Appearance and the two Motion streams and insert it in Fused Method.\par
            $ $BF$ \gets Fused($BA$,$BM$)$.\par
            \BState Take the Output of Fused Appearance and the two Motion streams and insert it in COA Track Method.\par
            $ $BTrack$ \gets COA($BF$)$.\par
           \State \Return \texttt{$ BF $ , $ CA $ ,$BTrack$}.\par
           \State $Frame\leftarrow Frame+1$\par
    \EndWhile
    \end{algorithmic}
    \end{algorithm}

\subsection{Detection Network}
In this paper, we use YOLO \cite{Redmon2017} a new approach to object detection, for bounding box prediction and classification. YOLO treats object detection in each frame as a regression problem and outputs spatially separated bounding boxes and associated class
probabilities. YOLO \cite{Redmon2017} ] is faster than region-proposal-based methods because it uses a single network to predict the bounding boxes and class probabilities in a single-pass evaluation. We have trained three YOLO detection networks: one to process the RGB frames, another to process the $x$ components of the motion vectors, and a third to process the $y$ components of the motion vectors. Additionally, we have trained a detection network for optical flow frame extraction based on \cite{Brox2011} following the approach in \cite{Zhang2016} and have used the weights of this network for transfer learning for extension to the motion vector frames.

\subsection{Motion Vector}
This paper use a motion-based detection network to improve the scores of the appearance-based detection network. Motion vectors, which already exist in a compressed video bit-stream, provide sufficient information to indicate the approximate location of a target object. A motion vector is comparable to an optical flow and can be used for object tracking and person detection. They are both two-dimensional vectors used to identify the motion of corresponding pixels in two continuous frames. In contrast to optical flows, however, motion vectors are used commonly in different video coding standards (e.g., HEVC and H.264). Thus, they are available from the compressed video stream and can be directly obtained at nearly no computational cost. This advantage makes motion vectors an powerful substitute for optical flows when attempting to obtain efficient action analysis.

In \cite{Zhang2016} a deep neural network framework based on motion vectors was presented to provide a solution to the main difficulty arising from the noise and inaccurate block-wise motion information offered by motion vector images. Motion vectors contain only block-level information and suffer from noisy and inaccurate motion information. Thus, training a CNN on motion vectors with high accuracy is difficult. Experiments demonstrate that directly using motion vectors in place of optical flow information will lead to accuracy degradations of 7\%, 10\% and 26\% accuracy on UCF101-24. Our aim is to take advantage of the real-time processing enabled by motion vectors to develop a model that is suitable for implementation in challenging environments, such as IoT systems, while still achieving a high detection accuracy comparable to that achieved using the optical flow approach. Driven by this motivation, we have applied the transfer learning approach \cite{Zhang2016} to design multiple methods of leveraging the rich and fine grained features that can be learned by an optical flow based network to improve our motion vector based network. These methods can be recognized as transferring the learned knowledge in the optical flow domain to the motion vector domain.

\subsection{Fusion of Appearance and Motion}
In the two-stream model, the motion features are processed separately and are then fused with the appearance stream. The fusion step is an important component of the two-stream model. Researchers have used various fusion methods, such as mean, max and multiplicative fusion. We consider two of these methods: mean fusion and max fusion. Algorithm \ref{alg:MulitFused} shows how our fusion algorithm works in detail.

For the motion stream, we first treat the Y motion as the basis stream and compare the X motion to it as follows: Let BMVX denote the X-motion-based detection box with the maximum overlap with a given Y-motion-based detection box, denoted by
BMVY. If this maximum overlap, quantified in terms of the intersection over union (IOU), is greater than a given threshold $t = 0:3$, we calculate the fused box as follows:

\begin{equation}
  \begin{split}
B_{MotionFused} = (B_{MVY} + B_{MVX})/2
 \end{split}
 \label{Eq:MotionFused}
\end{equation}

The second approach is to let BMVY denote the Y-motion-based detection box with the maximum overlap with a given X-motion-based detection box, BMVX. Similarly, if this maximum overlap, quantified in terms of the IOU, is greater than a given threshold $t = 0:3$, we again calculate the fused box using equation \ref{Eq:MotionFused}.\par

To extract the final bBounding box, let $B_{MotionFused}$  denote the fused motion-based detection box with the maximum overlap with a given appearance-based detection box, denoted by $B_{Appearance}$. If this maximum overlap, quantified in terms of the IoU, is greater than a given threshold $t = 0:3$, we calculate the final fused box as shown in equation \ref{Eq:FinalMFused} below.
This fusion method takes the box from each stream and returns the average of both the appearance and motion streams.

\begin{equation}
  \begin{split}
B_{Final} = (B_{MotionFused} + B_{Appearance)})/2
 \end{split}
 \label{Eq:FinalMFused}
\end{equation}

As an alternative, the max fusion method as shown in equation \ref{Eq:MaxFused} takes the box from each stream and returns the maximum between the appearance and motion streams. In this method, to ensure that the max fusion result will take advantage of the complementary contributions between the two streams, we take the maximum between  the boxes at each point, not the maximum area.

\begin{equation}
  \begin{split}
Fs = max(B_{MotionFused},B_{Appearance)}
 \end{split}
  \label{Eq:MaxFused}
\end{equation}

Algorithm \ref{alg:MulitFused} sshows the sequence followed in our approach to fuse the different streams to reach the required high accuracy. Here, we attempt to increase the accuracy by fusing the different bounding boxes identified from the different streams.

\begin{algorithm*}[!ht]
    \caption{Fused Multi-Stream}\label{alg:MulitFused}
    \hspace*{\algorithmicindent} \textbf{Input} :   \(\text{$threshold=.5$} \rightarrow \text{Intial threshold}\)\\
    \hspace*{\algorithmicindent} \textbf{Input} :   \(\text{$BA$      } \rightarrow \text{Boxes From Apperance Stream}\)\\
    \hspace*{\algorithmicindent} \textbf{Input} :   \(\text{$BMVX$       } \rightarrow \text{Boxes From Motion Vector X Stream}\)\\
    \hspace*{\algorithmicindent} \textbf{Input} :   \(\text{$BMVY$       } \rightarrow \text{Boxes From Motion Vector Y Stream}\)\\
    \hspace*{\algorithmicindent} \textbf{Input} :   \(\text{$BM$       } \rightarrow \text{Boxes From Fused Motion Stream}\)\\
    \hspace*{\algorithmicindent} \textbf{Input} : \(\text{$MList = 0$} \rightarrow \text{Boxes Founding in $BM$}\)\\
    \hspace*{\algorithmicindent} \textbf{Input} : \(\text{$LA$} \rightarrow \text{Number of boxes in $BA$ }\)\\
    \hspace*{\algorithmicindent} \textbf{Input} : \(\text{$LMVX $} \rightarrow \text{Number of boxes in $BMVX$ }\)\\
    \hspace*{\algorithmicindent} \textbf{Input} : \(\text{$LMVY $} \rightarrow \text{Number of boxes in $BMVY$ }\)\\
    \hspace*{\algorithmicindent} \textbf{Input} : \(\text{$LM $} \rightarrow \text{Number of boxes in $BM$ }\)\\
    \hspace*{\algorithmicindent} \textbf{Input} : \(\text{$FM$}\rightarrow \text{Fused Method $Mean$ or $Max$ }\)\\
    \hspace*{\algorithmicindent} \textbf{Output} :\(\text{$BF$} \rightarrow \text{Final Boxes From Two Fused Stream}\)\\
    \begin{algorithmic}[1]
   \If {$LMVY == \textit{0}$}
    \State \texttt{$ \textit{No Boxes Found in Current Motion Vector Frame} $}
    \Else
    \For{\texttt{< $i=1 \textit{ to } LMVX $ >}}
        \If {$LMVY > \textit{0}$}
            \For{\texttt{< $J=1 \textit{ to } BM $ >}}
                \State \texttt{$ \textit{ Compute } IOU(BMVXi,BMVYj)$}
            \If {$IOU(BMVXi,BMVYj) > threshold $} %\ref{EQ:IOU}
                \State \texttt{$ \textit{ ADD } BMVYj \textit{ to } MList $}
            \EndIf
                \State $j \gets j+1$
            \EndFor
            \State \texttt{$ M = arg max(MList)$}
            \If {$M_{MV} > 0 $}
                    \State  \texttt{$ \textit{Add } Mean(BMVXi,M_{MV}) \textit{ to } BM $ } \ref{Eq:MotionFused}
            \Else
                     \State  \texttt{$ \textit{Add } BMVXi \textit{ to } BM $ }

            \EndIf
        \Else
        \State  \texttt{$ \textit{Add } BAi \textit{ to } BF $ }
        \EndIf

       % \else
        %No suitable box is found  Take the target location in frame
        %\EndElse

        %\State $i \gets i+1$
    \EndFor
    \EndIf

    \If {$LM == \textit{0}$}
    \State \texttt{$ \textit{No Boxes Found in Current Frame} $}
    \Else
    \For{\texttt{< $i=1 \textit{ to } LA $ >}}
        \If {$LM > \textit{0}$}
            \For{\texttt{< $J=1 \textit{ to } BM $ >}}
                \State \texttt{$ \textit{ Compute } IOU(BAi,BMj)$}
            \If {$IOU(BAi,BMj) > threshold $} %\ref{EQ:IOU}
                \State \texttt{$ \textit{ ADD } BMj \textit{ to } MList $}
            \EndIf
                \State $j \gets j+1$
            \EndFor
            \State \texttt{$ M = arg max(MList)$}
            \If {$M > 0 $}
                \If {$FM == \textit{Max}$}
                    \State  \texttt{$ \textit{Add } Max(BAi,M) \textit{ to } BF $ } \ref{Eq:MaxFused}
                \Else
                    \State  \texttt{$ \textit{Add } Mean(BAi,M) \textit{ to } BF $ } \ref{Eq:FinalMFused}
                \EndIf
            \Else
                \State  \texttt{$ \textit{Add } BAi \textit{ to } BF $ }
            \EndIf
        \Else
        \State  \texttt{$ \textit{Add } BAi \textit{ to } BF $ }
        \EndIf
       % \else
%        No suitable box is found  Take the target location in frame
%        \EndElse

        \State $i \gets i+1$
    \EndFor
    \EndIf
    \State \Return \texttt{$ BF $ }.
    \end{algorithmic}
    \end{algorithm*}

\subsection{Object Tracking Using the COA}

Object tracking is described as estimating of the trajectory of a target object through a video frame sequence. The ultimate goal of human tracking is to automatically initialize and track all humans in a scene. In this paper, we apply the COA \cite{Pierezan2018} for human  actor tracking. Each actor being tracked is represented by a rectangular window centered on the middle point of the target person. These rectangular windows are obtained from our action localization and detection network after the fusion of the different streams to obtain the most accurate location of each action and the corresponding actor. Each such location is represented in the form P=( x , y , w , h), where the (x , y) coordinates are the centre of the box and the (w, h) parameters are width and height of the box, respectively. Then we are applying the COA To search this four dimensional feature space. Because the motion of an object is usually continuous, we can suppose the target is moving to some new location that is near its location in the previous frame. First, a set of coyotes (individuals/solutions) is initialized in a sub-search space around the target object, which is defined as A = ( $x_{s}$, $y_{s}$ , $w_{s}$ , $h_{s}$ ), where the ($x_{s}$, $y_{s}$) coordinates represent the central point of the sub-search space and the ( $w_{s}$ , $h_{s}$ ) parameters are its width and height, respectively, and are equal to (w, h), i.e., width and height of the bounding box of the target object. The search space is updated in each frame in accordance with the following equations:

\begin{equation}
	\label{EQ:SSX}
	X^t_{s} = X^{t-1}+ V^{t-1}_{X}
\end{equation}

\begin{equation}
	\label{EQ:SSY}
	Y^t_{s} = Y^{t-}+ V^{t-1}_{Y}
\end{equation}

Where$ V^{t-1}_{X}$ and $V^{t-1}_{Y}$ are represent target object’s velocities on horizontal and vertical axis in last frame and obtain by eq.8 ; then, the image extracted from each search window is used to compare with the target object to evaluate the fitness of that Coyote.\par

A fitness function is used for quantification of the strength of each coyote. The fitness function used for evaluation of the fitness value (F), which measures the similarity of each candidate coyote with respect to the target object and is used to determine, for each coyote, whether its new condition is better than its previous one according to equation 10. This equation represents the personal evaluation of each coyote, which will drive the coyote to move towards the best position that it has found so far. Thus, the velocity in the search space is dynamically adjusted according to the experience of each coyote or the collective experience of its pack. \par

There are many approaches for calculating the fitness value, most of which are based on histogram comparisons. In this paper, our aim is to develop a real-time tracking approach, meaning that there is a need to reduce the computational burden and processing time. Therefore, we use an alternative approach \cite{Liu.G2016} to measure the similarity between each coyote and the target object based on the $l2$ distance, which can be geometrically interpreted as the Euclidean distance between two vectors. This distance takes the following form:

\begin{equation}
	\label{EQ:DL2}
	Y_{l2}(P_{s},P_{t})= \sqrt{\sum_{i}\sum_{j}(P_{s(ij)}-P_{t(ij)})}
\end{equation}
Where \emph{Ps} denotes the pixel of the image captured by the search window of a Coyote and \emph{Pt} denotes the pixel of the target image.
To find the fitness value F, the \emph{dl2} value is divided by where \emph{Am} represents the area of the search window for the $m_th$ target object. And is then normalized to the range of 0 to 1. A value of 0 indicates a complete mismatch, and a perfect match corresponds to a value of
1. That is, the higher the fitness value, the more similar the coyote's search window to the target object. the fitness function for a coyote is defined as follows:

\begin{equation}
	\label{EQ:fitness}
	F= f(\frac{d_{l2}}{A_{m}})
\end{equation}

In the tracking of a moving object, Occlusion is one of the most challenging faced, in which the target may be covered by some background feature or may even disappears from the field of view and then re-enter the screen from an uncertain location. Using the technique presented in \cite{Liu.G2016} for detecting and handling occlusion, we build a histogram-based target model from the beginning of the video sequence to the current time point, which we will replace with the target search window image. The target model is recalled and compared against the current best-adapted coyote, which will be the next alpha in the pack. If the likelihood of similarity is lower than some predefined occlusion detection threshold, denoted by FT, then the target will be marked to be lost and the search space is to be extended.

\section{Experimental results}
In this experimental section, we introduce first the dataset used in our experiments and then analyse the experimental results for the proposed model of human action detection and tracking.

\subsection{Dataset}
We evaluate our model on the the UCF-101-24 \cite{Soomro2012}. UCF-101-24 is a subset of UCF-101 \cite{Soomro2012}, one of the largest and most diversified and challenging action datasets. UCF-101 includes total number of 101 action classes which have divided into five types: Human-Object Interaction, Body-Motion, Sports , and Playing Musical Instruments. Each video includes only one category of action, it may contain multiple action instances of the same action class (up to 12 in a video) , with different temporal and spatial boundaries. UCF101-24 is a subset consisting of 24 of the 101 classes and includes spatio-temporal localization annotation in the form of bounding box annotations for human targets, released as part of the THUMOS-2015 challenge.

\subsubsection{Experimental Settings}
The goal is for our model to be suitable for implementation in challenging environments such as IoT environments; thus, the training of our model was a one time one-time process. We used a machine with a Tesla k80 GPU, an Intel Xenon CPU and 14 gigabytes of RAM. We considered the RGB frames first and trained the appearance model; this phase took three days to complete. Then, we used the method introduced in \cite{Brox2011} to generate optical flow frames from the RGB frames and used these frames to train our motion stream detection network, and finally, we used the motion vector frames generated by the video compression tools and applied transfer learning to extend the trained optical flow weights to a motion-vector-based motion model.\par

\subsection{Human Action Detection and Localization Results}

Experiments were designed to study the effectiveness of the appearance stream, the motion stream and the two streams after fusion. For the fusion of the motion streams using equation \ref{Eq:MotionFused}. we found that the first approach achieves an accuracy of up to 23.25\% at $\delta$=0.2 , whereas the accuracy of the second approach reaches 25.93\% at $\delta$ =0.2; thus, we chose to use the second approach to obtain all further results. When we use equation \ref{Eq:FinalMFused} to fuse the motion and appearance streams, an accuracy of is achieved 72.12\% at $\delta$=0.2, and an accuracy of 62.74\% at $\delta$=0.5.

\begin{figure*}[htbp]
  \centering
  \includegraphics[width=6in, height=3in]{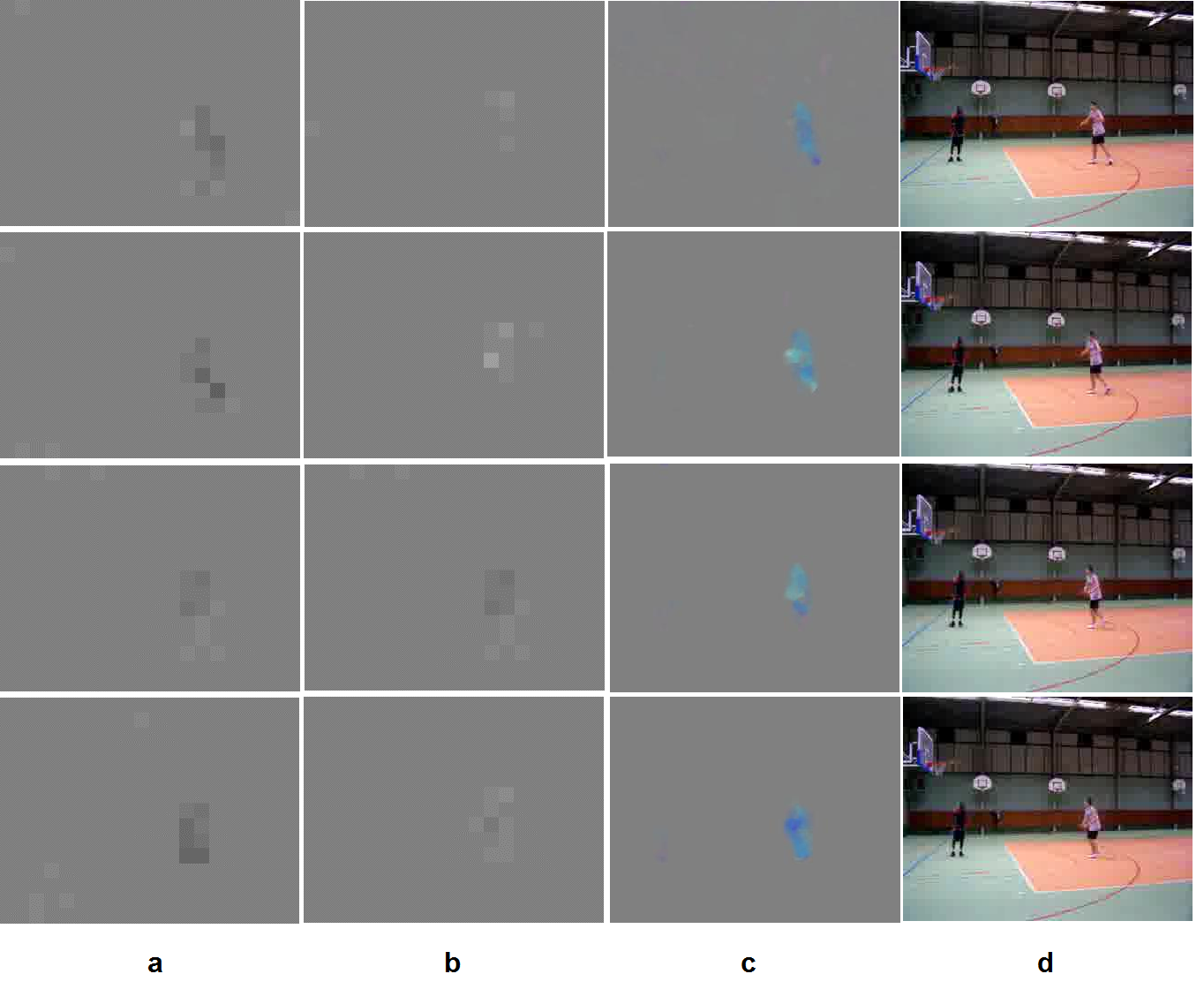}
  \caption{(a, b and d) The input to our stream from the video decoder, (c) is the optical flow frame using in training.}
  \label{fig:StreamInput}
\end{figure*}

Figure \ref{fig:StreamInput}(a, b and d) shows the inputs to our model obtained from the video decoder, which converts a video stream into motion vector frames and RGB frames. Figure (2-a) shows the X-component motion vector information, Figure (2-b) shows the Y-component motion vector information, and Figure (2-c) shows the corresponding optical flow frame, which is not an input to our network but rather is the output of \cite{Brox2004} . We trained and tested an initial motion detection network on the optical flow information and then used the trained weights to perform transfer learning on the motion vector dataset. Figure (2-d) shows the RGB frame used to detect the appearance information for our action model.

Table \ref{tab:table1Multi} shows the class-specific precision (AP) in \% achieved for the videos in each action category of UCF-101-24 by using the detection networks based on the appearance stream and the two fused motion vector streams separately and by using the multi-stream (appearance and motion) fusion model.\par

The results were generated using a threshold of $\delta$ =0.20. For 12 of the 24 action classes, our appearance + motion fusion technique yields the best APs. The appearance-based detection network alone achieves the best APs for 11 of the 24 classes. \par

%==========================================================================
%

\begin{table*}[htbp]
\caption{Detection results (video APs in \%) on UCF-101-24.}
\label{tab:table1Multi}
\begin{tabular*}{\linewidth}{lllllll}
  \hline
\\[-1em]
\textbf{Actions}           & \textbf{Basketball} & \textbf{BasketballDunk}  & \textbf{Biking}         & \textbf{CliffDiving}   & \textbf{CricketBowling}    & \textbf{Diving}            \\\hline \\[-1em]
\textbf{Saha et al. \cite{Saha2016}}        & 39.6 &49.7& 66.9& 73.2& 14.1& 93.6   \\\hline \\[-1em]
\textbf{Singh et al. \cite{Singh2017}}        & 42.0&64.6& 73.7& 75.2 &41.5& 100.0   \\\hline \\[-1em]
\textbf{appearance}        & 68.77      &  75.15   &   80.73     &       52.32  &  42.94  &   95.80   \\
\textbf{motion}      &3.05          &45.89&18.47&20.087&29.80&22.54 \\
\textbf{appearance+motion (Mean)} &68.83&79.52 &79.60 &52.57&43.31&95.62\\\hline \\[-1em]

\textbf{Actions}                    & \textbf{Fencing}    & \textbf{FloorGymnastics}& \textbf{GolfSwing}  & \textbf{HorseRiding}     & \textbf{IceDancing}     & \textbf{LongJump}        \\\hline \\[-1em]
\textbf{Saha et al. \cite{Saha2016}}        & 85.9 &99.8   &  68.3 &94.1 &63.1 &57.2   \\\hline \\[-1em]
\textbf{Singh et al. \cite{Singh2017}}        & 86.5&97.9   &  62.1 &96.0& 77.6& 69.7   \\\hline \\[-1em]
\textbf{appearance}        &   88.87   &      86.14     &    73.81 & 99.14 &  53.4        &    47.11       \\
\textbf{motion}            & 45.43&47.85&7.60&75.477&51.82&30.16\\
\textbf{appearance+motion (Mean)} &88.25&86.11&73.78&99.14&54.08&47.41\\\hline \\[-1em]
\textbf{Actions}           & \textbf{PoleVault}         & \textbf{RopeClimbing}       & \textbf{SalsaSpin}      & \textbf{SkateBoarding}  & \textbf{Skiing}     & \textbf{Skijet}         \\\hline \\[-1em]
\textbf{Saha et al. \cite{Saha2016}}        & 75.1&89.6&31.1&85.1&79.6&96.1   \\\hline \\[-1em]
\textbf{Singh et al. \cite{Singh2017}}        & 76.1 &96.1& 22.2& 87.4 &81.0&87.1   \\\hline \\[-1em]
\textbf{appearance}        &    51.97   &   79.62   &   56.17     &71.58      &     64.62  & 95.48         \\
\textbf{motion}            &  15.44&18.81&32.0&19.99&8.56&20.25\\
\textbf{appearance+motion (Mean)} &52.69&79.55&55.69&71.41&64.47&95.36\\\hline \\[-1em]
\textbf{Actions}         & \textbf{SoccerJuggling} & \textbf{Surfing} & \textbf{TennisSwing}       & \textbf{TrampolineJumping}&
\textbf{VolleyballSpiking} & \textbf{WalkingWithDog}    \\\hline \\[-1em]
\textbf{Saha et al. \cite{Saha2016}}        & 89.1&63.2&33.6&52.7&20.9&75.6   \\\hline \\[-1em]
\textbf{Singh et al. \cite{Singh2017}}        & 82.4&62.1&37.4&59.4&21.7&85.1 \\\hline \\[-1em]
\textbf{appearance}        &    78.32  &    91.27 &  68.06   & 65.43  &57.27  &82.57 \\
\textbf{motion}            &37.30&49.37&2.78&24.68&3.90&7.22\\
\textbf{appearance+motion (Mean)} & 78.31&91.10&68.34&65.78&57.98&82.33\\\hline \\[-1em]
\end{tabular*}
\end{table*}

%
%==========================================================================

Table \ref{tab:table2Multi} presents the results we obtained on UCF-101-24, First, to prove the strength of our model in localizing and predicting the actions of different actors in different environments, we compare our model with the top offline and online competitors \cite{Singh2017}, \cite{Peng2016}, \cite{Saha2016},\cite{Peng2016} and \cite{Weinzaepfel2015} in terms of detection performance.

All of the results reported in table II for other competitors were obtained using the offline approach, although \cite{Singh2017} also presented an online model with a speed of 28 fps and a mean AP (mAP) of 70.2\% . All of the results of our model were obtained online, at a speed of 57 fps for our appearance-stream-only method and at a speed of more than 52 fps for multi-stream fusion.

%==========================================================================
%
\begin{table}[htbp]
 \caption{Action localization results (mAP) on UCF101-24 dataset.}
 \label{tab:table2Multi}
\begin{tabular}{lllll}
  \hline
\\[-1em]
\textbf{IoU threshold $\delta$ }    & \textbf{0.2} & \textbf{0.5} & \textbf{0.75} & \textbf{0.5:0.95} \\
\\[-1em]
  \hline
\\[-1em]
Weinzaepfel et al. \cite{Weinzaepfel2015} & 46.8         & –            & –             & –                 \\
Peng and Schmid \cite{Peng2016}    & 73.5         & 32.1         & 02.7          & 07.3              \\
Saha et al. \cite{Saha2016}        & 66.6         & 36.4         & 07.9          & 14.4              \\
Singh et al. \cite{Singh2017}        & 73.5         & 46.3         & 15.0          & 20.4              \\
\\[-1em]
  \hline
\\[-1em]
Ours-Appearance (A)         &  71.94            &   61.18           &               &                   \\
Ours-Motions (M)            &  25.93     &              &               &                   \\
Ours-A + M (Mean-Fuse)                  & 72.12         &     62.74         &               &                   \\
\\[-1em]
  \hline
\end{tabular}
\end{table}
%
%==========================================================================

Figure \ref{fig:MultiResult} shows examples of the human action localization results obtained on UCF101-24 \cite{Soomro2012}. Each row represents one test video clip from UCF-101. As shown in the top rows, our model can successfully localize more than one action per frame with high accuracy and speed. The results show the superiority of our model in detecting multiple actions in the same frame and in localizing and predicting small targets, as in the surfing examples.

\begin{figure*}[htbp]
  \centering
  \includegraphics[width=6in, height=3in]{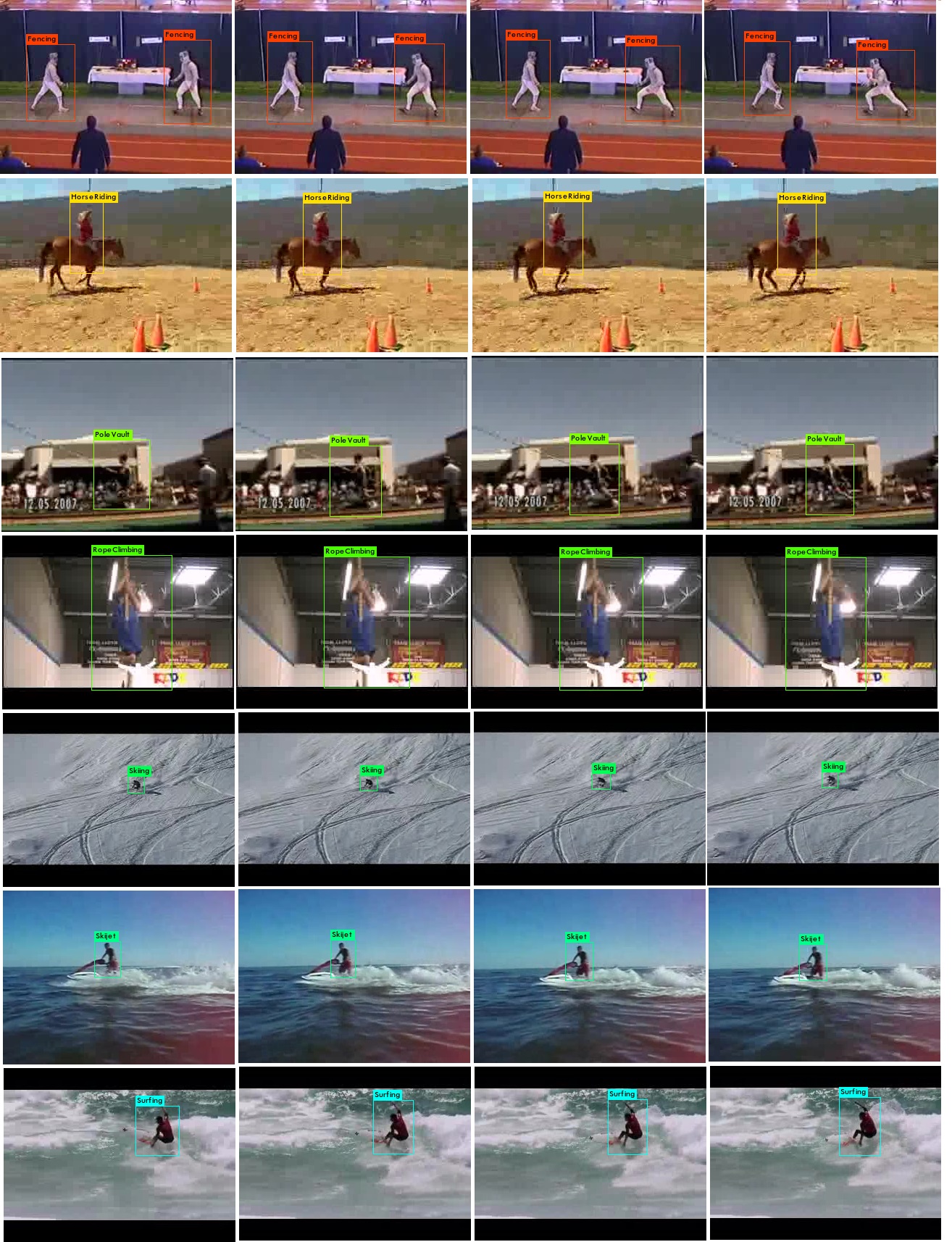}
  \caption{Human action localization results on UCF101-24.}
  \label{fig:MultiResult}
\end{figure*}

Table \ref{tab:table3Multi} compare our detection speed to those reported by Singh et al. \cite{Singh2017} and Saha et al. \cite{Saha2016}.  Our model has a detection speed of 57 fps when using RGB frames only and a speed of 52 fps when using motion vectors frames.\par
%==========================================================================
%

\begin{table}[htbp]
\caption{Test time detection speed.}
 \label{tab:table3Multi}
\begin{tabular*}{\linewidth}{lcccc}
  \hline
\textbf{Framework modules}           & \textbf{A}  & \textbf{A+M} & \multicolumn{1}{l}{\textbf{A+M}} \cite{Singh2017}\\ \hline
\textbf{Flow computation (ms)}       & -                       & 2              & 7                                            \\
\textbf{Detection network time (ms)} & 14.9                     & 14.9           & 21.8                                         \\
\textbf{Overall speed (fps )}        & \multicolumn{1}{l}{\textbf{57}} & \textbf{52}    & \textbf{34.7}        \\\hline
\textbf{ $\star$ not include time to generate tube}        &   &    &      \\\hline
\end{tabular*}
\end{table}
%
%==========================================================================
\subsection{Results for Real-Time Human Tracking Using the COA}
We test our model using ucf-24 \cite{Soomro2012} datasets. A value of $FT =0.90$ was used in the COA to determine whether to conduct a global search. A lower value of FT corresponds to a more global search and more iterations are generally required and coyotes for covering a larger search space while maintaining accuracy of tracking. A higher value of FT will lead in a less global search and consequently less iterations and coyotes, That can improve the tracking speed. However, it will sometimes result to failure in handling occlusion problems.\par

Figure \ref{fig:MultiResult} shows the tracking results for our model which has a speed of up to 20 fps when only the CPU is used for processing. The model has the ability to track more than one actor by using multiple swarms, one for each actor.

\begin{figure*}[htbp]
  \centering
  \includegraphics[width=6in, height=3in]{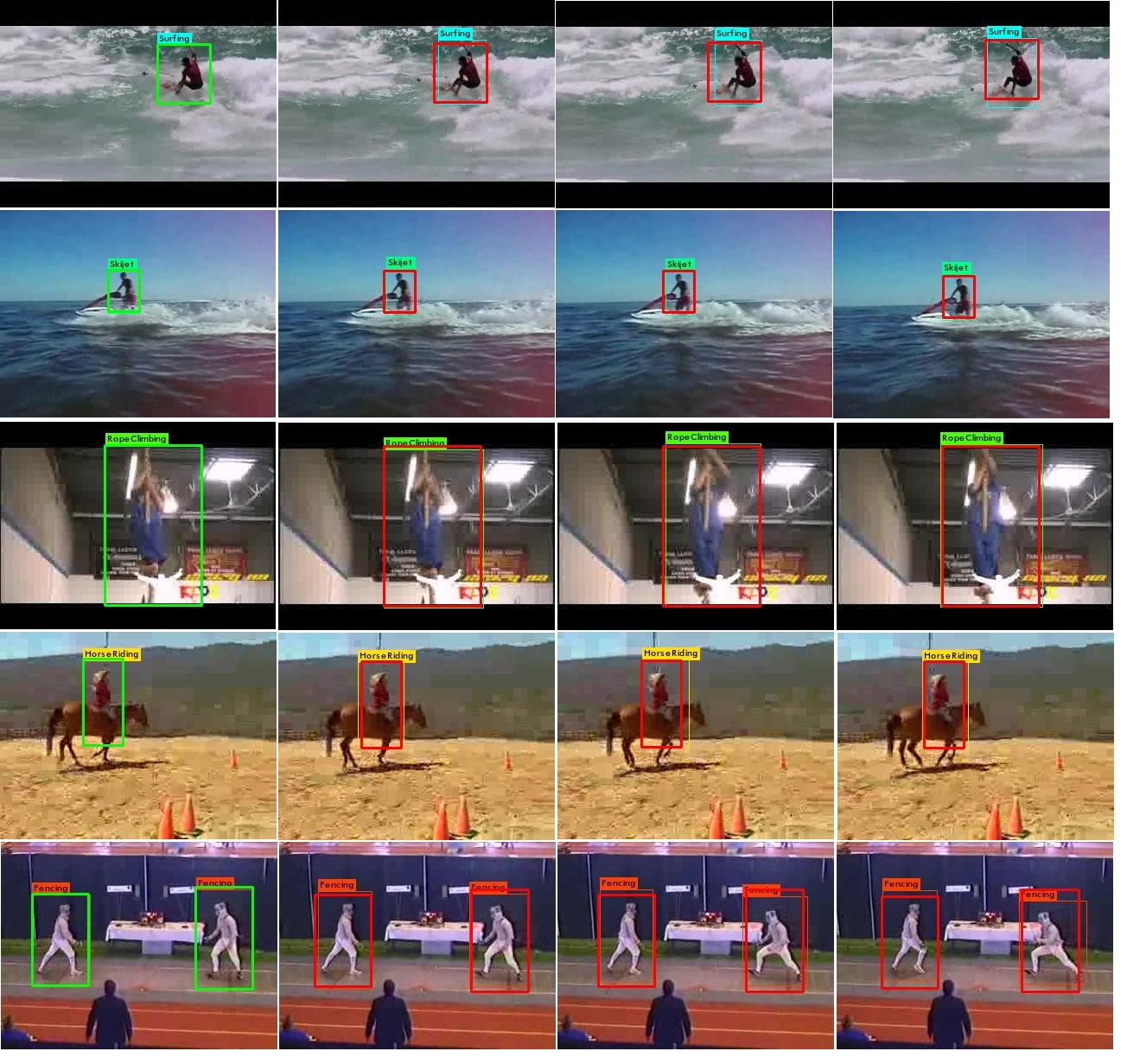}
  \caption{Tracking Result for our Human Action Tracking and localization Model.}
  \label{fig:MultiResult}
\end{figure*}

\subsection{Tracking Results Obtained Using the COA vs. YOLO}

%\begin{figure*}[htbp]
%  \centering
%  \includegraphics[width=\textwidth]{Images/Fitness.png}
%  \caption{The fitness function for COA tracking model.}
%  \label{fig:Fitness}
%\end{figure*}

\begin{figure*}[htbp]
  \centering
  \includegraphics[width=6in, height=3in]{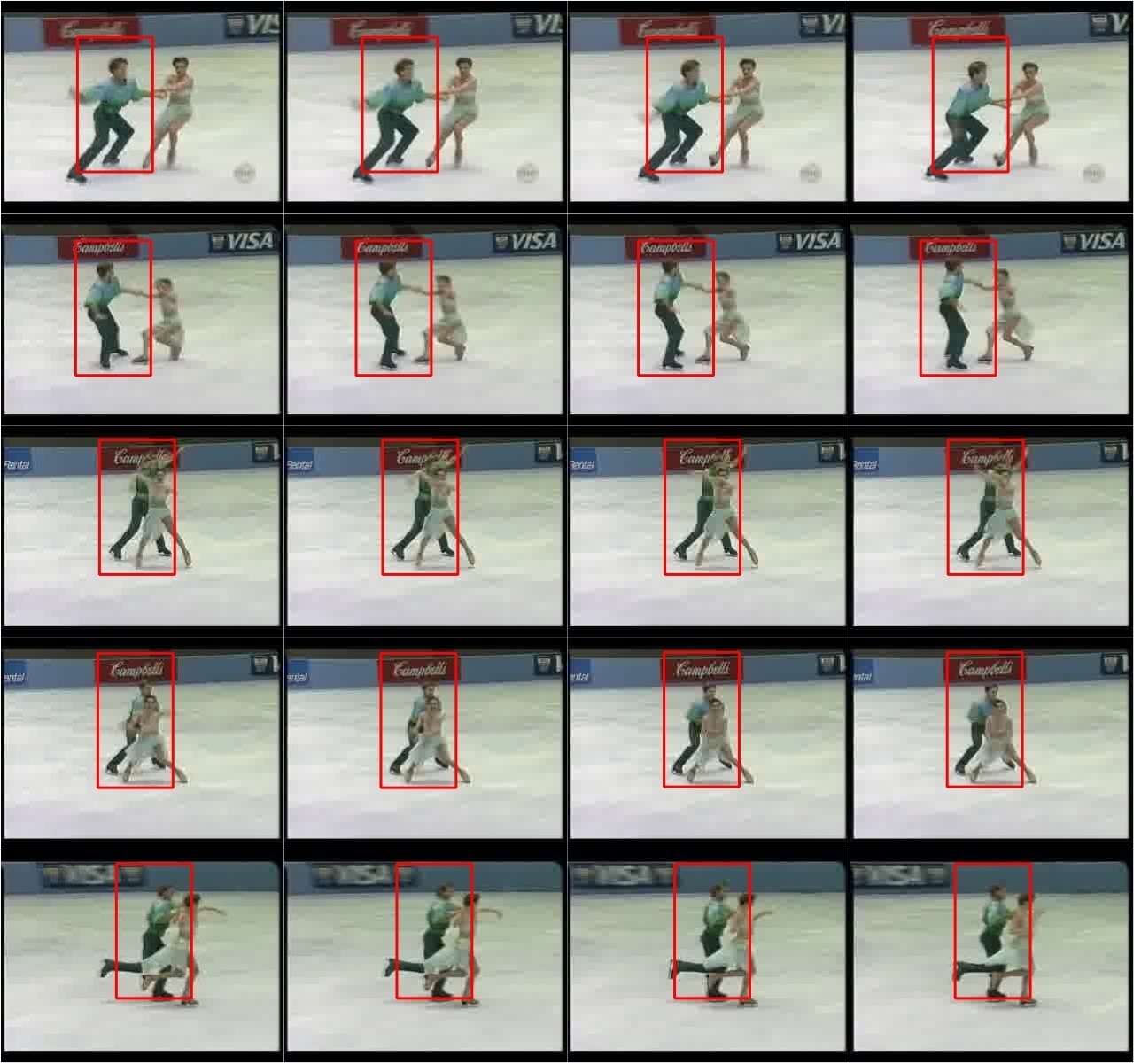}
  \caption{COA Tracking Result for Human.}
  \label{fig:COATracking}
\end{figure*}

In this section, we show the advantages and disadvantages of using a COA-based tracker as a tracking model by comparing it with a tracking model based on YOLO \cite{Redmon2017}, which is based on an approach that we call tracking by detection because the detection network attempts to detect the target in each frame. As seen in Figure \ref{fig:COATracking}, the COA-based tracker achieves full accuracy and success in tracking the target object across different frames (248 out of 248 frames) and can also distinguish the target actor from other actors that collide with it, as shown in rows 3 and 4. However, the COA has a drawback that arises from the adopted fitness function: the tracker requires a clear target object, and it may fail in detecting the intended target and instead select another target that has a close similarity with the true target. Moreover, in the COA-based approach, an additional tracker is needed to track each additional object to be tracked (one tracker for each target actor). The COA has a target fitness function with a minimum threshold of = $90\%$. Every value represents the best solution in each target frame.

\begin{figure*}[htbp]
  \centering
  \includegraphics[width=6in, height=3in]{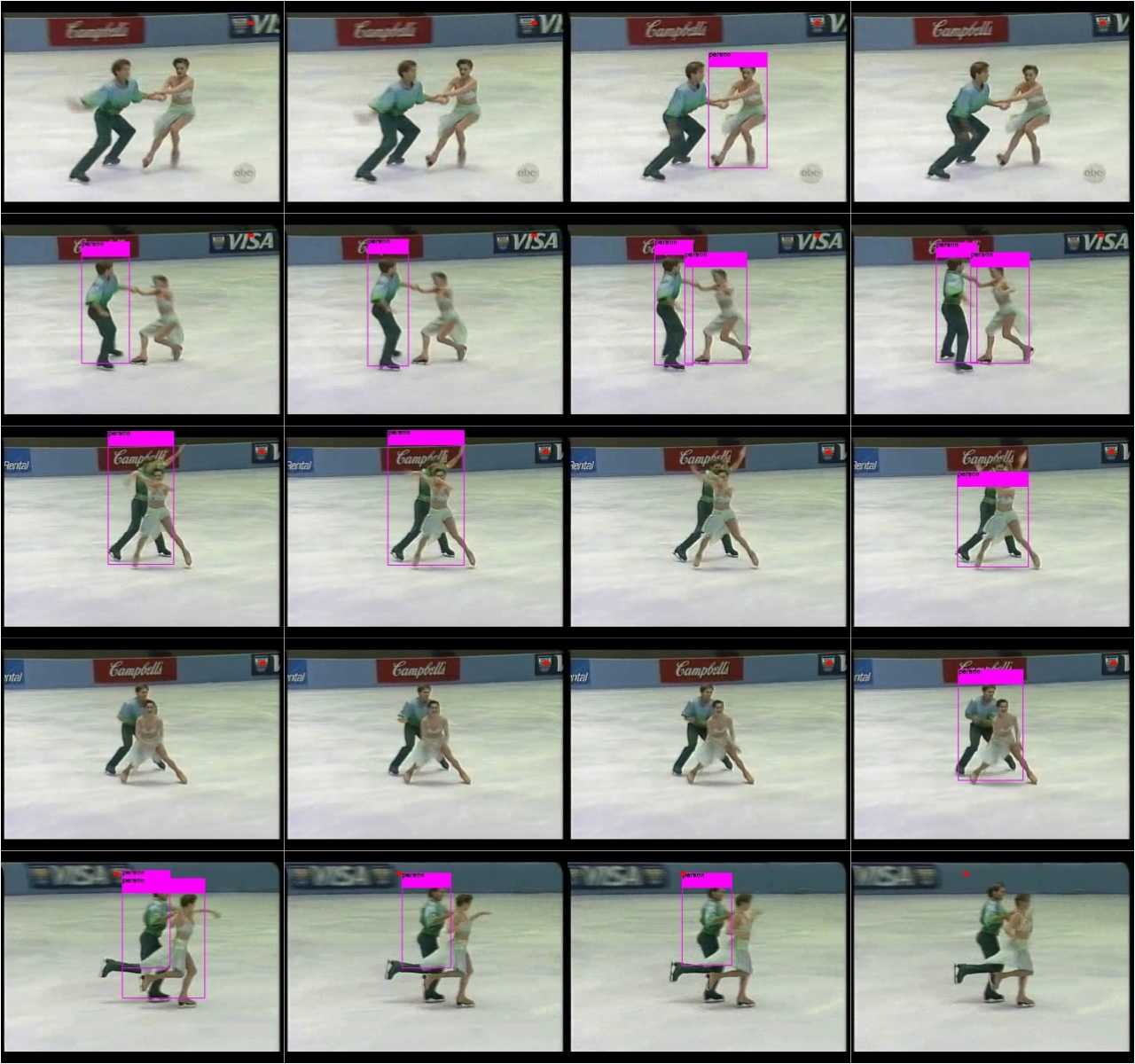}
  \caption{YoloV2 Tracking Result for  Human.}
  \label{fig:YoloV2Tracking}
\end{figure*}

As seen in Figure \ref{fig:YoloV2Tracking}, YOLOv2 \cite{Redmon2017} shows lower accuracy than the COA-based tracker in detecting the actor in every frame of the selected video. YOLO must detect every target actor while treating every frame as a new source; therefore, YOLO alone cannot identify that a given detected object is the same object as an object detected in the previous frame. YOLOv2  achieves good accuracy and success in detecting the target object in different frames (157 out of 248 frames). \par

\begin{figure*}[htbp]
  \centering

  \includegraphics[width=6in, height=3in]{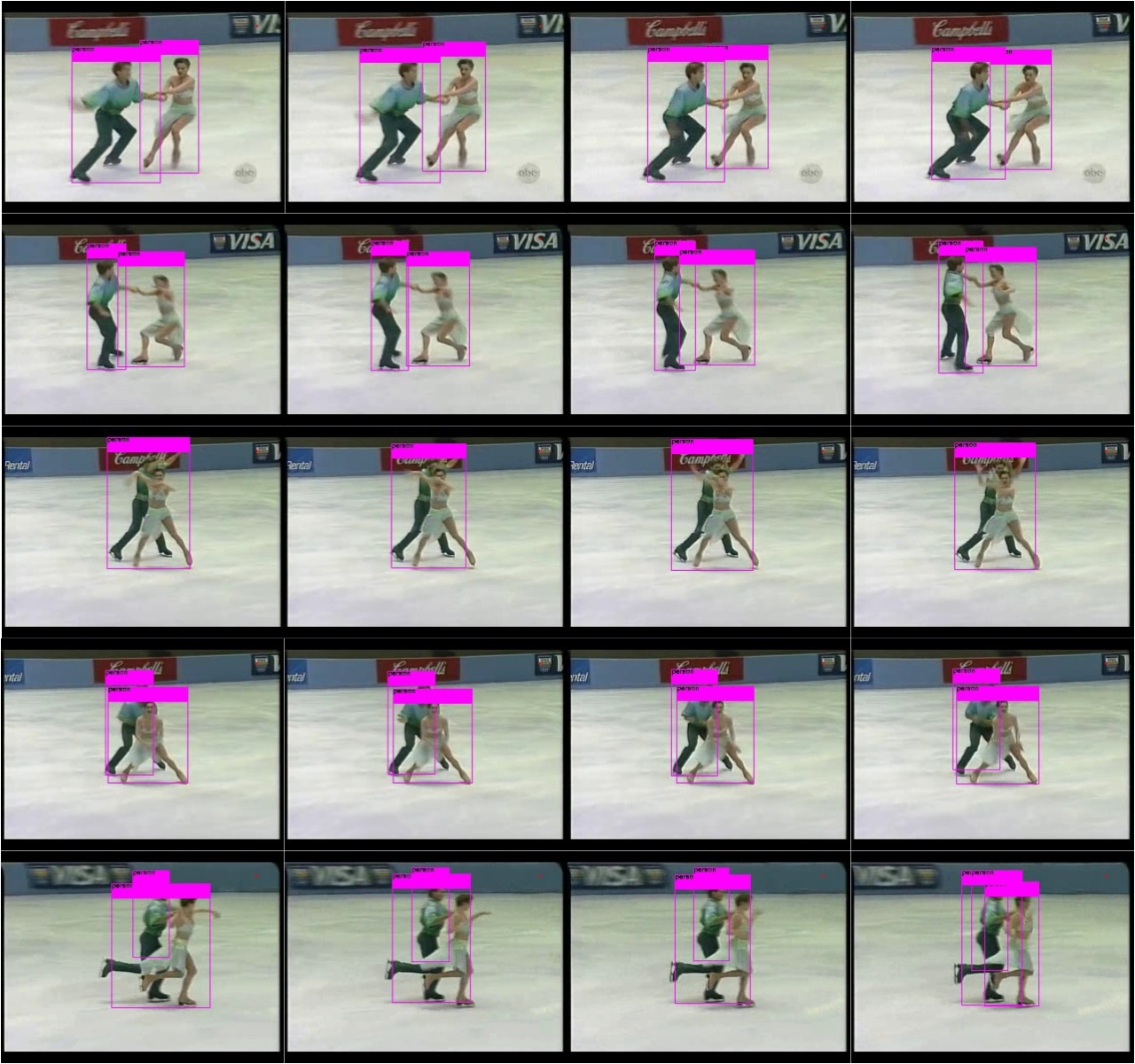}
  \caption{YoloV3 Tracking Result for  Human.}
  \label{fig:YoloV3Tracking}
\end{figure*}

As seen in Figure \ref{fig:YoloV3Tracking}, YOLOv3 \cite{Redmon2018} shows higher accuracy than YOLOv2 \cite{Redmon2017}, successfully detecting the actor in every frame of the selected
video. YOLOv3 also detects every target actor. YOLOv3 can distinguish objects from other objects that collide with them, as shown in rows 3 and 4; however, because it still treats every frame as a new source, YOLOv3 alone still cannot identify that a detected object is the same object as an object detected in the previous frame. \par

Tracking actors in real time requires both high speed and good accuracy. A COA-based tracker achieves good performance in little processing time but also has certain disadvantages. YOLOv3 achieves good accuracy in detecting target objects but treats each frame as a new source of data. In future work, we will attempt to combine YOLOv3 with the COA to develop a high-accuracy real-time tracker that can detect target objects efficiently and cope with the collision problem. Note that YOLOv3 achieves full accuracy and success in detecting the target object in different frames (248 out of 248 frames).\par

\section{Conclusion and Future Work}

Real-time action localization refers to the task of simultaneously localizing actions and identifying their classes from video stream. It is a challenging problem that requires expensive features that are difficult or impossible to extract due to the real-time processing requirements. The localization and classification of actions must be performed even before the actions are fully observed. In this work, we have introduced a fast and accurate model that can achieve action detection with an accuracy of 72\% (mAP)  and a speed of  58\% fps. Our model can simultaneously detect multiple actions. It also produces good results for action prediction in videos using information available in the current frame only. The proposed tracking model based on the COA has the ability to track more than one actor by using multiple swarms, one for each actor.\par

In future work, we will consider a multi-camera scenario in which it is necessary to detect an action and track the corresponding actor across different cameras as the actor moves from one place to another. We will also consider the tracking of multiple objects. Furthermore, we will combine the COA with YOLOv3 to develop an efficient tracker that can cope with collisions while maintaining a high detection accuracy rate.

%\section*{Acknowledgment}

\ifCLASSOPTIONcaptionsoff
  \newpage
\fi

\end{document}